\documentclass[11pt]{article}
\usepackage{geometry}
\usepackage{coling2020}
\usepackage{times}
\usepackage{url}
\usepackage{latexsym}
\usepackage{microtype}
\usepackage{graphicx}
\usepackage{url}
\usepackage{amsmath}
\usepackage{verbatim}

\usepackage[T1]{fontenc}
\usepackage[latin9]{inputenc}
\usepackage[table]{xcolor}
\usepackage{collcell}
\usepackage{hhline}
\usepackage{pgf}
\usepackage{multirow}

\colingfinalcopy 

\title{UPB at SemEval-2020 Task 6: Pretrained Language Models for Definition Extraction}

\author{Andrei-Marius Avram\textsuperscript{1,2}, Dumitru-Clementin Cercel\textsuperscript{1}, Costin-Gabriel Chiru\textsuperscript{1} \\
  University Politehnica of Bucharest, Faculty of Automatic Control and Computers\textsuperscript{1} \\
  Research Institute for Artificial Intelligence, Romanian Academy\textsuperscript{2} \\
  {\tt avram.andreimarius@gmail.com}, \\
  \tt \{clementin.cercel, costin.chiru\}@cs.pub.ro\\}

\date{}

\begin{document}
\maketitle
\begin{abstract}
This work presents our contribution in the context of the 6th task of SemEval-2020: Extracting Definitions from Free Text in Textbooks (DeftEval). This competition consists of three subtasks with different levels of granularity: (1) classification of  sentences as definitional or non-definitional, (2) labeling of definitional sentences, and (3) relation classification. We use various pretrained language models (i.e., BERT, XLNet, RoBERTa, SciBERT, and ALBERT) to solve each of the three subtasks of the competition. Specifically, for each language model variant, we experiment by both freezing its weights and fine-tuning them. We also explore a multi-task architecture that was trained to jointly predict the outputs for the second and the third subtasks.
Our best performing model evaluated on the DeftEval dataset obtains the 32nd place for the first subtask and the 37th place for the second subtask. The code is available for further research at: \url{https://github.com/avramandrei/DeftEval}
.
\end{abstract}

\section{Introduction}

\blfootnote{This work is licensed under a Creative Commons Attribution 4.0 International Licence. Licence details: http://creativecommons.org/licenses/by/4.0/.
}
Definition extraction from text is a challenging research task, addressed by numerous researchers in the area of natural language processing (NLP). Factual question-answering systems are possible applications that can benefit from the results of this task \cite{zhang2009automatic}. As a response to this challenge, \newcite{spala2019deft} introduced the Definition Extraction from Texts (DEFT) corpus, a human-annotated English dataset that contains multi-domain (e.g., biology, sociology, physics) term-definition pairs from two types of documents, free (i.e., Textbook) and semi-structured text (i.e., Contracts), as opposed to the domain-specific WCL dataset \cite{navigli2010learning}.
In addition, a shared task was proposed by SemEval-2020 which was aimed at evaluating the performance of each participant system on three subtasks defined for the DEFT corpus. The three subtasks are the following:

\textit{\textbf{Subtask 1}: Given a labeled  dataset of sentences, this subtask is to build a classifier capable of distinguishing between a sentence containing both a definition and the defined term
or not.} We provide two examples of sentences that contain a definition, from Biology and Physics domains, respectively:
\begin{itemize}
    \item  \textit{The metabolome} \textbf{is} \textit{the complete set of metabolites that are related to the genetic makeup of an organism}.
    \item \textit{Polarization} \textbf{is} \textit{the separation of charges in an object that remains neutral}.
\end{itemize}

\textit{\textbf{Subtask 2}: Given a dataset of tokenized sentences, we aim to label each token with one of the following classes: Term, Alias-Term, Referential-Term, Definition, Referential-Definition, or Qualifier.} The meaning of each token can be found in the corpus description paper \cite{spala2019deft}. 

\textit{\textbf{Subtask 3}: Given a dataset of tokenized sentences and the tag id of each token, our goal is to predict, for each token, the type of relationship  it had with another token, and also the tag id of the token it had a relation with.} This is a classical relation extraction task \cite{zeng2018large} and the relations that must be extracted are: Direct-defines, Indirect-defines, Refers-to, AKA, and Supplements.

Previous approaches \cite{klavans2001evaluation,fahmi2006learning,zhang2009automatic} of recognizing definition sentences mainly focused on the use of linguistic clues (e.g., "is", "means", "are", "a", or "()").
However, these studies fail to classify sentences containing definitions, where the linguistic clues are non-existent.
In recent years, neural network-oriented solutions are another line of research in order to capture definitions from the text.
\newcite{anke2018syntactically} have proposed an architecture that relies on two models, convolutional neural network \cite{fukushima:neocognitronbc} and bidirectional long short-term memory network \cite{hochreiter:1997}. 
Recently, \newcite{veyseh2019joint} combines more advanced deep learning techniques by leveraging graph convolutional neural networks with both syntactic and semantic information.
However, the existing methods for definition extraction cannot benefit from language pretrained models \cite{lan2019albert}. Motivated by their recent performances in more NLP tasks, we employ these models for solving the previously mentioned subtasks.

Subtask 2 is related to many NLP tasks that can be formalized as a sequence labeling problem, such as
part-of-speech tagging  \cite{liu2018empower},
named entity recognition \cite{ma2016end,dumitrescu2019introducing},
metaphor detection  \cite{wu2018neural},
emphasis selection  \cite{shirani2019learning},
keyphrase extraction \cite{alzaidy2019bi},
fragment-level propaganda detection  \cite{da2019findings},
complex word identification  \cite{gooding2019complex},
language identification  \cite{mave2018language},
or 
cyber attacks detection in server logs \cite{ghimes2018character}. 

For the third subtask, we did not take into consideration the fact that we were given the set of correct tags of the second subtask (which tremendously help in the relation classification) and, thus, our approaches obtained poor results on both development and test datasets, so this subtask will not be discussed in the rest of the paper.

Our main contributions can be summarized as follows:
\begin{itemize}
    \item We explore various pretrained language models and we depict their results for the first and the second subtasks, respectively.
    \item As has been shown on other corpora \cite{peters2019tune}, we find that fine-tuning the weights of the pretrained language models on the DEFT corpus gives a boost in performance, as opposed to freezing them. 
    \item Additionally, we investigate a RoBERTa model \cite{liu2019roberta}  within a multi-task architecture that jointly learns the outputs of the second and the third subtask. However, we report only its performance for the second subtask for the reason presented above.
\end{itemize}

\section{Neural Architectures}

\subsection{Pretrained Language Models}

All pretrained language models presented below use the Transformer encoder \cite{vaswani2017attention} to produce contextualized embeddings. The Transformer is a self-attention mechanism that can capture long-distance dependencies between its inputs.
For each language model, we use the implementation that is publicly available in the
HuggingFace repository\footnote{https://github.com/huggingface/transformers}.

\subsubsection{BERT}

Bidirectional Encoder Representations from Transformers (BERT) \cite{devlin2018bert} was bidirectionally trained using two strategies: (1) Masked Language Modeling (MLM), and (2) Next Sentence Prediction (NSP), on both the BooksCorpus \cite{zhu2015aligning} with a dimensionality of 800M words and the English Wikipedia with a dimensionality of 500M words. It has improved the existing results on the General Language Understanding Evaluation (GLUE) dataset \cite{wang2018glue} by 7\%.

\subsubsection{XLNet}

XLNet \cite{yang2019xlnet} trains with a new objective, called Permutation Language Modeling (PLM), that instead of predicting the tokens in a sequential order in the same way as a traditional autoregressive solution, it predicts the tokens in a random order. Moreover, aside from using PLM, XLNet relies on Transformer XL \cite{dai2019transformer}, a variation of the transformer structure that can capture a longer context through recurrence, as its base architecture. XLNet surpass BERT on a series of NLP tasks, and set the new state-of-the-art
result on the GLUE dataset with an average F1-score of 88.4\%.

\subsubsection{RoBERTa}

Later on, the Robustly optimized BERT  pretraining approach (RoBERTa) \cite{liu2019roberta} uses the MLM strategy of BERT, but removes the NSP objective. Moreover, the model was trained with a much larger batch size and learning rate, on a much larger dataset and showed that the training procedure can significantly improve the performance of BERT on a variety of NLP tasks. In particular, with a F1-score of 88.5\%, RoBERTa reached the top position on the GLUE leaderboard, outperforming the previous leader, XLNet. When RoBERTa was released, it obtained new state-of-the-art results on the GLUE dataset, improving the performance of BERT by 6.4\% and over the previous leader by 0.1\%.

\subsubsection{SciBERT}

Science BERT (SciBERT) \cite{beltagy2019scibert} is a pretrained language model based on BERT.
As opposed to BERT, the novelty here is that SciBERT 
was trained on large multi-domain corpora of scientific publications to improve performance on domain-aware NLP tasks. Experimental results show that SciBERT significantly surpassed BERT on biomedical, computer science, and other scientific domains. 

\subsubsection{ALBERT}

A Lite BERT (ALBERT) \cite{lan2019albert} is a system that has fewer parameters than the classical BERT, but still maintains a high performance. The contributions of ALBERT consist
in two key approaches of parameter reduction: factorized embedding parameterization and cross parameter sharing.
In addition, ALBERT uses for training the sentence-order prediction instead of NSP. ALBERT obtained an average F1-score of 89.4\% on the GLUE dataset, pushing the state-of-the-art by 0.6\%.

\subsection{Conditional Random Fields}

The most common method for treating a sequence labeling task is the Conditional Random Field (CRF) model \cite{lafferty2001conditional}.
As was mentioned by \newcite{alzaidy2019bi}, for a sequence $x$ of input words and another sequence $y$ of output tags, CRF works by constructing a conditional probability distribution in the following manner: 
\begin{equation}
    p(y|x;W, b) \propto exp\left(\sum_{i=1}^{n} {W^T_{y_{i-1}, y_i}x_i + b_{y_{i-1}, y_i}}\right)
\end{equation}

\noindent
where the parameters $W_{y_{i-1}, y_i}$ and $b_{y_{i-1}, y_i}$ are called the weight matrix and the bias, respectively.

To estimate the parameters $W$ and $b$, we perform a maximization of the log-likelihood function:
\begin{equation}
    L(W, b) = \sum_{j=1}^{N} {log(y^{(j)}|x^{(j)};W,b)}
\end{equation}    
    
Once the CRF is trained, we use the Viterbi algorithm \cite{forney1973viterbi} to find the most probable sequence among all possible tag sequences.

\subsection{Approaches based on Pretrained Language Models}

For the first subtask, we add a two-layered feed-forward neural network with 512 neurons in each layer on top of the \texttt{[CLS]} contextualized embedding (as proposed in the original paper of BERT for the single sentence classification tasks \cite{devlin2018bert}) that maps this embedding to a scalar. By applying a sigmoid function to this mapping, we obtain a trainable scalar that represents the probability of a sentence to contain a definition. 


For the second subtask, we also map the contextualized embeddings generated by each pretrained language model in a lower-dimensional space using a two-layered feed-forward neural network. Then, we use these mappings to train a CRF model that learns to predict the most probable sequence of labels for a given input.  The main problem that we encountered in this subtask was that we needed to apply a special tokenization, namely Byte Pair Encoding (BPE), 
to train the language models,  that was different than the one used to create the corpus. To mitigate this issue, we reconstructed the sentence from its tokens and splited it again in BPE subtokens. Then, to map the subtokens back to the original tokens, we employed a character matching algorithm that is similar to the one used by spacy-transformers\footnote{More details about the algorithm can be found at the Tokenization Alignment section from this repository: \url{https://github.com/explosion/spacy-transformers}.}. 

\begin{figure*}
    \centering
    \includegraphics[width=0.7\textwidth]{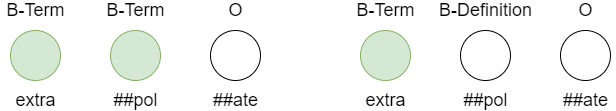}
    \caption{Label mismatch for the word "extrapolate" - green represents the chosen label. Left: the model predicts for two of the three subtokens the label B-TERM, so it will become the label of the word. Right: there is no label majority, so the first subtoken is chosen.}
    \label{fig:label_missmatch}
\end{figure*}

The BPE tokenization also introduced a problem at inference because the predicted labels for the subtokens of a word might not match, so a label for the whole word could not be  inferred directly. To solve this problem, we took the label of the majority or, if the labels were equally distributed, we selected the label of the first subtoken. Figure \ref{fig:label_missmatch} depicts the proposed solution for the problem of label mismatch for the word "extrapolate".

\subsection{Multi-task Learning Approach}

In this work, we also experimented with a language model that jointly learned to predict the tags, the tag ids, and the relations. Figure 2 depicts the architecture of the multi-task learning setting. 
To create the framework, we projected the contextualized embeddings generated by RoBERTa in three vectors, representing the outputs for the subtasks 2 and 3, respectively.

The approach on tag prediction subtask in the multi-task context is identical to its single subtask counterpart. To predict the tags ids, we consider that the maximum number of possible tags in a paragraph is 10 and that the id of the tag is given by its position in this context. Once we identify all tag ids, we predict the corresponding relations.

\begin{equation}
    L(y_{tag}, y_{id}, y_{rel}, \hat{y}_{tag},\hat{y}_{id}, \hat{y}_{rel}) = 
    \lambda_1L_1(y_{tag}, \hat{y}_{tag}) 
    + \lambda_2L_2(y_{id}, \hat{y}_{id}) 
    + \lambda_3L_3(y_{rel}, \hat{y}_{rel})
\end{equation}

\begin{figure*}
    \centering
    \includegraphics[width=0.55\textwidth]{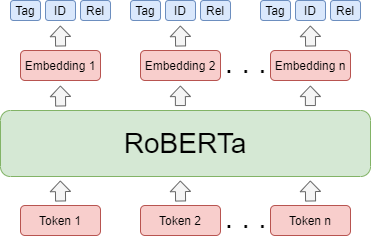}
    \caption{Our multi-task learning framework for the subtasks 2 and 3, respectively.}
    \label{fig:my_label}
\end{figure*}

The learning  objective of the multi-task method is to predict the outputs for the three subtasks by minimizing the following multi-task loss function: 

\noindent
where $y_{tag}, y_{id}, y_{rel}$ are the true labels for tags, tag ids and relations, respectively, while $\hat{y}_{tag},\hat{y}_{id}, \hat{y}_{rel}$ are the corresponding predictions.
Also, $\lambda_1, \lambda_2, \lambda_3$ and $L_1, L_2, L_3$ are the weights and the individual loss functions, respectively.

\section{Performance Evaluation}

\subsection{Dataset and Preprocessing}

The DeftEval dataset contains imbalanced classes for all the subtasks considered. For example, the first subtask has 11,090 sentences that do not contain a definition and 5,569 that contain. 
In order to handle this issue, we balance the classes by doubling the number of sentences that contain a definition, obtaining a new total of 11,138 positive samples. Moreover, the second task has highly imbalanced classes, ranging from 93,204 for the Definition tag to 256 for the Referential-Term tag. To balance the classes in this case, we oversampled each sentence that contains an under-represented tag by a factor inversely proportional to their number\footnote{We note that by using this method, the number of over-represented tags increases as well, but by a much smaller factor.}. The number of the initial and final entries for each tag (i.e., before and after applying the oversampling technique), along with the corresponding multiplication factor are depicted in Table \ref{tab:oversample}. As one can note, the final number of tags is not equal to the initial number of tags multiplied by its factor. This is because other tags were affected when we oversampled a sentence for a certain tag, with a particular multiplication factor\footnote{When we oversampled a sentence, we considered only the tag that had the highest multiplication factor in that sentence.}.

\begin{table}[]
    \centering
    \begin{tabular}{lccc}
        \hline
        
        \textbf{Tag type} & \textbf{No. of initial samples} & \textbf{No. of final samples} & \textbf{Multiplication factor}  \\
        \hline \hline 
        Definition & 93,204 & 115,190 & 1 \\
        Term & 16,162 & 16,439 & 1 \\
        Alias-Term & 1,586 & 6,444 & 4 \\
        Qualifier & 1,207 & 5,136 & 4 \\
        Ref-Definition &  969 & 7,824 & 8 \\
        Ref-Term & 256 & 4,096 & 16 \\
        \hline
    \end{tabular}
    \caption{Statistics of each tag class before and after applying the oversampling technique.}
    \label{tab:oversample}
\end{table}

The preprocessing step consists of removing the artifacts that could interfere with the language model representation of the sentences. For the first task, we replace the URLs and equations with two special tokens, \texttt{<url>} and \texttt{<equation>}, respectively, during the fine-tuning process, and remove them while freezing the language model weights. 
We also eliminate the artifacts that came from the text formatting\footnote{The text formatting artifacts usually came from expressions, like \texttt{size 12 \{ \}}.} and the spaces before punctuation. For the second subtask, we discard the characters that were not recognized by the language models and that could break down the tag-subtoken matching process, like the "º" character or Greek letters. We also replace the accented characters with their corresponding unaccented version.  
\subsection{Experimental Settings}

For training purposes, we employ the Adam optimizer \cite{kingma2014adam} with a learning rate of $2e-5$ for both the frozen and fine-tuned versions. We train each language model for 100 epochs with a batch size of 16 and we save only the language model that obtained the highest performance on the development dataset. We decided not to use the early stopping setting because we observed that the models can significantly improve the efficiency of our results even after a long period of stagnation.

The feed-forward layer, which  is placed on top of each language model to project the contextualized embeddings in the output space, has a hidden layer of size 512. Thus, a more complex family of functions can be learned by the system. We regularize the hidden layer with a high 80\% dropout for the fine-tuned versions in order to make their learning slower and more robust, and with a 20\% dropout for the frozen versions to avoid underfitting.

Due to the computational constraints, we adopt only the base version of all the language models tested in the current work. Furthermore, we use only the cased version of each language model when there is the case. Finally, we select the cross-entropy loss function for the multi-task architecture and we set all the weights $\lambda_1, \lambda_2, \lambda_3$ equal to $0.33.$ 

\subsection{Results and Analysis}

As mentioned above, we conducted experiments with a total of five pretrained language models, including BERT, XLNet, RoBERTa, SciBERT, and ALBERT. More specifically, we use each language model by freezing and fine-tuning its weights. We also experiment with a multi-task architecture that is a joint learning technique to predict the outputs for both the second and the third subtask. 
Table 2 reports the evaluation metrics, Macro-Precision, Recall, and F1-scores, respectively, on both development and test datasets.

\begin{table*}[]
    \centering
    \begin{tabular}{lcccccc}
          & \multicolumn{3}{c}{\textbf{Dev}} & \multicolumn{3}{c}{\textbf{Test}} \\
          \hline
          \multicolumn{1}{c}{\textbf{Model}} & \textbf{Precision} & \textbf{Recall} & \textbf{F1-score} & \textbf{Precision} & \textbf{Recall} &  \textbf{F1-score} \\
          \hline\hline
          Frozen BERT & 76.0 & 75.1 & 75.5 & - & - & - \\
          Frozen RoBERTa & 74.1 & 74.4 & 74.3 & - & - & - \\
          Frozen SciBERT & 71.7 & \textbf{80.6} & \textbf{75.9} & - & - & - \\
          Frozen XLNet & 66.8 & 70.8 & 68.7 & - & - & - \\
          Frozen ALBERT & \textbf{77.2} & 69.3 & 73.0 & - & - & - \\
          \hline
          Fine-tuned BERT  & 78.4 & 84.1 & 81.2 & - & - & -\\
          Fine-tuned RoBERTa & 78.2 & \textbf{84.5} & \textbf{81.3} & 75.0 & 80.6 & \textbf{77.7}\\
          Fine-tuned SciBERT & \textbf{79.4} & 79.7 & 79.6 & - & - & -\\
          Fine-tuned XLNet & 75.5 & 85.2 & 80.1 & - & - & -\\
          Fine-tuned ALBERT & 69.5 & 85.9 & 76.8 & - & - & -\\
          \hline
          The winning team & - & - & - & - & -  & \textbf{87.7} \\
          \hline
          \\
    \end{tabular}
    
    \begin{tabular}{lcccccc}
          & \multicolumn{3}{c}{\textbf{Dev}} & \multicolumn{3}{c}{\textbf{Test}} \\
          \hline
          \multicolumn{1}{c}{\textbf{Model}} & \textbf{Precision} & \textbf{Recall} & \textbf{F1-score} & \textbf{Precision} & \textbf{Recall} &  \textbf{F1-score} \\
          \hline\hline
          Frozen BERT+CRF & 27.1 & \textbf{39.8} & 26.1 & - & - & - \\
          Frozen RoBERTa+CRF & \textbf{32.7} & 27.7 & 22.6 & - & - & - \\
          Frozen SciBERT+CRF & 29.0 & 37.5 & 26.2 & - & - & - \\
          Frozen XLNet+CRF & 29.6 & 33.4 & \textbf{26.8} & - & - & - \\
          Frozen Multi-task Learning & 4.0 & 8.9 & 8.0 & - & - & - \\
          \hline
          Fine-tuned BERT+CRF  & \textbf{47.9} & 51.7 & 45.6 & - & - & - \\
          Fine-tuned RoBERTa+CRF & 41.4 & \textbf{66.4} & \textbf{46.0} & 39.4 & 55.6 & \textbf{43.9} \\
          Fine-tuned SciBERT+CRF & 46.7 & 46.6 & 41.7 & - & - & - \\
          Fine-tuned XLNet+CRF & 33.0 & 58.5 & 39.2 & - & - & -\\
          Fine-tuned Multi-task Learning & 25.7 & 25.2 & 25.5 & - & - & - \\
          \hline
          The winning team & - & - & - & - & -  & \textbf{84.7} \\
          \hline
    \end{tabular}
    
    \caption{Comparison of performance on both the development and test datasets for the first subtask (up) and the second subtask (down), respectively. The frozen versions denote that the weights of the respective language model were not updated during the training phase, while the fine-tuned versions denote that their weights were updated during the training process.}
    \label{tab:metrics}
\end{table*}

It can be observed from the two tables that fine-tuning the weights of the language models offers a high boost in performance. That is, the results show an improvement of up to 11.4\% on the development dataset for the first subtask (in case of XLNet) and up to 23.4\% on the development dataset for the second subtask (in case of RoBERTa). Moreover, the results on the development dataset also show that a fine-tuned RoBERTa is the best performing language model among all others, for both subtasks. Thereby, this was the only model submitted for evaluation, and it obtained a F1-score of 0.777 for the first subtask, and a macro F1-score of 0.439 for the second subtask, ranking 32nd and 37th on the leaderboard, respectively. 

\begin{figure}
    \centering
    \includegraphics[width=0.99\textwidth]{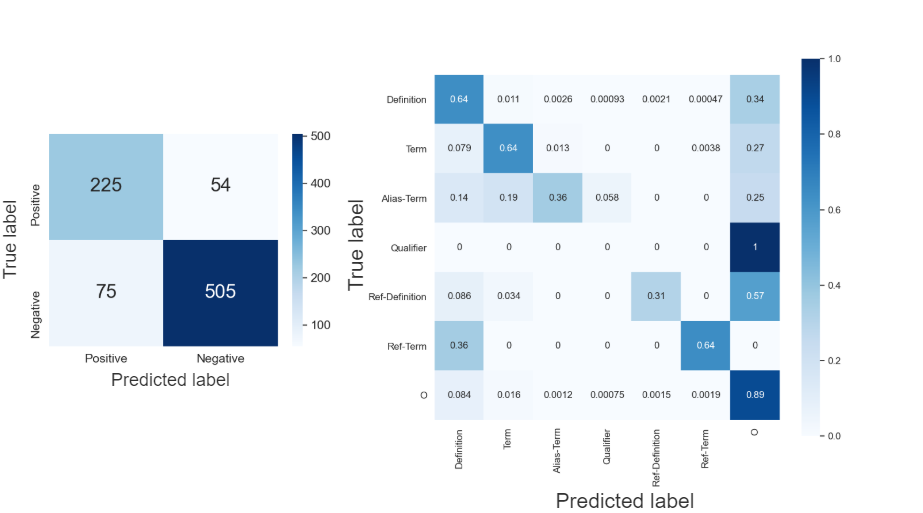}
    (a)  \qquad \qquad  \qquad \qquad \qquad  \qquad \qquad \qquad \qquad  (b) \qquad \qquad \qquad 
    \caption{Confusion matrix for (a) the first subtask using the fine-tunned RoBERTa model and (b) the second subtask using the fine-tunned RoBERTa+CRF model.}
    \label{fig:conf_matrix}
\end{figure}

Figure \ref{fig:conf_matrix} depicts the detailed confusion matrices of the submitted models for the first and the second subtasks on the evaluation dataset. In the case of the first subtask, we can observe that the model is slightly biased towards predicting positive labels, resulting in more false positives for the input sentences, which is somewhat expected, given the balancing method we have applied. To improve the visualization of the results, the confusion matrix for the second subtask was normalized along the true labels for the test set. It can be seen that the dominant tags, Definition and Term, and also the Ref-Term, the tag that was highly oversampled, were the least confused by the model, obtaining an accuracy of 64\%, which is almost double than the accuracy obtained by the rest of the tags. Moreover, it should be noted that most of the tags are misclassified by the model with the O tag, exception making the Ref-Term tag which is misclassified with the Definition tag, 36\% of the time.



\section{Conclusions and Future Developments}

In this paper, we have presented our solution for the SemEval-2020 Task 6: Extracting Definitions from Free Text in Textbooks (DeftEval). We evaluated different state-of-the-art language models both by freezing and fine-tuning their weights. We observed that the performance of all the selected models can be significantly improved by fine-tuning their weights. 
Through a series of experiments conducted on the development dataset, we showed that RoBERTa significantly outperforms other language models for the two subtasks. According to the official leaderboard, we obtained the 32nd place out of 56 submissions for the first subtask and the 37th place out of 51 submissions for the second subtask.
One possible direction for future work is to evaluate the  multi-task scenario using other language models.
We also consider that a larger annotated dataset together with the large variants of the pretrained language models could drastically improve definition extraction performance. Moreover, we believe that by using class weights instead of oversampling for the first subtask, one can mitigate the problems observed in its confusion matrix.


\bibliographystyle{coling}
\bibliography{coling}

\end{document}